\documentclass{amsart}
\usepackage{graphicx} 
\usepackage{amsfonts, amsmath, amssymb, amsthm}
\usepackage{tikz}
\usepackage{hyperref}
\usepackage{subcaption}
\usepackage{booktabs}
\usetikzlibrary{positioning}
\usepackage{bbold}
\makeatletter
\def\amsbb{\use@mathgroup \M@U \symAMSb}
\makeatother

\theoremstyle{definition}
\newtheorem{definition}{Definition}[section]
\newtheorem{example}{Example}[section]

\theoremstyle{remark}
\newtheorem*{remark}{Remark}

\newcommand{\AC}{\mathrm{AC}}
\newcommand{\Acc}{\mathrm{Acc}}
\newcommand{\Stb}{\mathrm{Stb}}
\newcommand{\ES}{\mathrm{ES}}
\newcommand{\E}{\amsbb{E}}
\newcommand{\bF}{\mathbf{F}}
\newcommand{\R}{\amsbb{R}}

\title[Beyond Accuracy]{Beyond Accuracy: A Stability-Aware Metric for Multi-Horizon Forecasting}

\author[Ma]{Chutian Ma}
\email{c.ma@causify.ai}
\author[Pomazkin]{Grigorii Pomazkin}
\email{g.pomazkin@causify.ai}
\author[Saggese]{Giacinto Paolo Saggese}
\email{gp@causify.ai}
\author[Smith]{Paul Smith}
\email{paul@causify.ai}

\begin{document}

\begin{abstract}
    Traditional time series forecasting methods optimize for accuracy alone. This objective neglects temporal consistency, in other words, how consistently a model predicts the same future event as the forecast origin changes. We introduce the forecast accuracy and coherence score (forecast AC score for short) for measuring the quality of probabilistic multi-horizon forecasts in a way that accounts for both multi-horizon accuracy and stability. Our score additionally allows user-specified weights to balance accuracy and consistency requirements. As an example application, we implement the score as a differentiable objective function for training seasonal auto-regressive integrated models and evaluate it on the M4 Hourly benchmark dataset. Results demonstrate consistent improvements over traditional maximum likelihood estimation. Regarding stability, the AC-optimized model generated out-of-sample forecasts with 15.8\% reduced variance over forecasts targeting the same timestamp. In terms of accuracy, the AC-optimized model achieved considerable improvements for medium-to-long-horizon forecasts. While one-step-ahead forecasts exhibited a 3.9\% increase in MSE, forecasts from horizon three onward experienced improved accuracy, with a peak improvement of approximately 6\% in MSE at horizons 9-12. These results indicate that our metric successfully trains models to produce more stable and accurate multi-step forecasts in exchange for a relatively small degradation in one-step-ahead performance.
\end{abstract}
\thanks{Code available at \url{https://github.com/causify-ai/beyond_accuracy}}
\maketitle
\tableofcontents

\section{Introduction}

We consider a setting where multi-horizon, probabilistic forecasts are issued at
regular intervals. For example, one may consider the chance of rain on any given
day over the next week in a particular locale. With the advance of each day, we
observe an outcome we previously did not have access to, and, at the outer reach
of our forecasting horizon, we record an initial forecast. For the days in
between the newly observed outcome and the limit of the forecasting horizon, we
record updated forecasts. These forecast updates typically reflect at least two
things:
\begin{enumerate}
    \item New information not previously available (e.g., new weather
    observations).
    \item The reduction (by one time step) of forecast horizon, and hence,
    \emph{ceteris paribus}, forecast uncertainty.
\end{enumerate}

\subsection{Stability of Forecast Revision over Time}
In the setting where the forecast horizon is fixed (e.g., next-day weather
forecasts), the problem of forecast accuracy is well-studied and typically
quantified using proper scoring rules. In contrast, the focus of our work is on
\emph{forecast stability}, which has received comparatively little attention.
To relate this concept back to our weather forecasting example, suppose that
today is a Sunday and we are considering the daily chance-of-rain forecasts for
the next seven days. Our initial forecast for \emph{next} Sunday's chance of
rain is a seven-day ahead forecast, at the outer reach of our forecast curve.
Common experience would recommend that we take this forecast less seriously
than, say, the next-day chance-of-rain forecast that we will have access to on
Saturday, the day before next Sunday. As the days advance from Sunday through
Saturday, we observe an updated chance-of-rain forecast for next Sunday, and by
the time we reach Sunday and observe the outcome, we will have accumulated seven
sequential probabilistic forecasts. In fact, the stability of forecasts have
been well studied in the weather forecasting domain. We will discuss more of the
relevant work in section \ref{sec:related_work}. In addition,
\cite{scotese1994forecast} studied the same concept in economics, showing that
forecasters at the Federal Reserve may intentionally ``smooth" forecasts and
sacrifice some one-step-ahead accuracy for temporal consistency. 

What we propose to capture is a measure of the \emph{stability}, or lack
thereof, of these sequential forecast updates. Through several examples below,
we illustrate that forecast stability is a meaningful concept to measure and can
play an important role in discriminating between two competing sets of
forecasts. All things being equal--a concept we will rigorously define--we would
prefer sequential forecasts that appear harmonious rather than conflicting.

Additionally, we propose incorporating forecast stability in a loss function
while training time series models, with the extent to which stability is
weighted tailored to the forecasting needs for the problem at hand.

\subsection{Time-based Forecast Weights for Decision Problems}
In addition to capturing forecast stability, we propose incorporating
problem-tailored forecast-horizon based weightings in forecast quality metrics.
Using horizon-based weights is motivated by the way multi-horizon forecasts may
be used in various decision-making problems. In other words, the importance of
time weighting is not necessarily inherent in the forecasting procedure itself,
but rather in how the forecasts are used. To return to our weather example, we
note that the decision about whether to carry an umbrella outside on Sunday may
be postponed to the last minute, assuming one has easy access to an umbrella. A
decision about whether to wash a car might be made based upon a one or two day
rain forecast. However, decisions around contingency planning for an outdoor
event may need to be made much further in advance. If such intended usage is
captured in a forecast quality metric, then we may better judge forecast quality
for a given decision problem. When we also have control over the way forecasts
are made, we may then in turn use this metric as part of the training
optimization procedure.

\subsection{Comprehensive Time Series Metrics}

In the sequel, we propose a comprehensive time series metric framework that is
both natural and flexible enough to capture a wide range of time series
properties desirable in practice for effectively utilizing probabilistic
multi-horizon forecasts in making decisions.

\begin{remark}[Desiderata]
The following properties are desirable of a metric, all else being equal:
\begin{enumerate}
\item The more accurate a forecast, the better.
\item Accuracy of forecasts immediately prior to a decision point are more
important than forecasts too far ahead of or after a decision point (note that a
decision point may be well in advance of the observed outcome).
\item The less revision in a forecast, the better.
\end{enumerate}
\end{remark}

In many applications, decisions must be made at some point between the forecast
origin and the target time. Forecasts issued long before the decision point may
require down-weighting or discounting, while those close to the decision should
receive greater emphasis. Introducing a weighting scheme along the forecast path
allows one to focus on accuracy and stability in the periods that matter most.

Stability is also critical when early action is possible. If the sequence of
forecasts for a target is sufficiently stable, then decision makers can act with
greater confidence on early forecasts. Conversely, highly volatile forecast
sequences may delay action or require more robust risk management.

Together, these considerations motivate a framework in which accuracy and
stability are quantified separately and then combined in a weighted metric-space
score, allowing both dimensions to inform probabilistic multi-step
decision-making.

\section{Related Work}\label{sec:related_work}
Forecast stability remains an understudied area relative to forecast accuracy,
and the existing literature has developed in parallel across domains including
meteorology, economics and operations research. We review the relevant work
along two dimensions: whether stability is assessed with or without reference to
eventual outcomes, and whether the framework applies to point forecasts or
probabilistic forecasts.
\subsection{Observation-Independent Stability Measures}
The earliest systematic treatment of forecast revision consistency appears in
the meteorological literature. Zsoter et al.\ \cite{zsoter2009jumpiness}
introduced the \emph{jumpiness index} to quantify inconsistency between
consecutive forecasts issued by the ECMWF and Met Office ensemble prediction
systems. Their index measures the normalized difference between successive
forecasts targeting the same valid time using the average standard deviation of
the two forecast fields as a scaling factor to enable comparison across forecast
steps and weather parameters. Building on this foundation, they introduced a
taxonomy of forecast jump patterns — the \emph{flip}, \emph{flip-flop}, and
\emph{flip-flop-flip} — to classify sequences of sign-alternating
inconsistencies of increasing persistence. Their empirical results established
an important baseline finding: forecast jumpiness and forecast error are only
weakly correlated, meaning that more consistent forecasts do not necessarily
have lower error. Zsoter et al.\ explicitly identified the extension of their
framework to probabilistic forecasts as an open direction, noting that the
jumpiness index should be generalized beyond ensemble means treated as scalar
quantities.
Griffiths et al.\ \cite{griffiths2019flipflop} subsequently proposed the
\emph{Flip-Flop Index}, which is defined as the total variation of a forecast
revision sequence minus the range of the sequence, averaged by sequence length.
Griffiths et al.\ \cite{griffiths2021circular} later extended this framework to
vector-valued forecasts via the \emph{Circular Flip-Flop Index}, allowing the
metric to evaluate forecasts such as wind direction.

In the applied forecasting literature, Pritularga and Kourentzes
\cite{pritularga2024} introduced \emph{congruence} as a measure of vertical
forecast stability, directly quantifying the variance of forecasts made at
different origins targeting the same time. Godahewa et al.\
\cite{godahewa2025stability} further studied vertical and horizontal stability
in the context of multi-horizon forecasting benchmarks, establishing terminology
and empirical baselines that the present work builds upon. Klee and Xia
\cite{klee2025stability} and Genov et al.\
\cite{genov2025switching} have examined stability from an operations research
perspective, with the latter demonstrating that in environments with
forecast-induced switching costs, more stable forecasts deliver measurably
better downstream performance even when raw accuracy is equivalent.

A recent contribution by Zanotti \cite{zanotti2025stability} introduces the
\emph{Multi-Quantile Change} (MQC) metric, which measures instability by
tracking changes across a finite set of predictive quantiles between successive
forecast origins. MQC represents a meaningful step toward probabilistic
stability assessment, and its model-agnostic, scale-free design shares the
practical orientation of our framework. However, because it operates on a finite
set of pre-specified quantile levels rather than the full predictive
distribution, MQC cannot capture all aspects of distributional change — in
particular, shifts in distributional shape that occur between or beyond the
chosen quantile levels are invisible to the metric.

\subsection{Observation-Dependent Stability Measures}
A philosophically distinct approach is taken by Ehret
\cite{ehret2010convergence}, who introduced the \emph{Convergence Index} to
evaluate whether a sequence of forecasts approaches the eventual observation
with decreasing lead time, without oscillating between over- and
underestimation. Rather than measuring revision jumps directly, the Convergence
Index counts divergence and oscillation events relative to the truth, weighted
inversely by lead time and normalized to the unit interval, where zero denotes
perfect convergence and one denotes incessant divergence and oscillation. A
tolerance band suppresses penalization of negligibly small fluctuations. This
observation-dependent perspective captures something the revision-consistency
family cannot: a forecast sequence may appear stable by the Flip-Flop Index
while systematically drifting away from the truth, whereas the Convergence Index
correctly rewards monotone convergence toward the outcome regardless of the
magnitude of the initial error. Ehret explicitly recommends combining the
Convergence Index with an absolute accuracy measure to obtain comprehensive
evaluation, acknowledging that the index is inherently relative and does not
assess absolute forecast quality in its own right.

\subsection{Positioning of the Forecast AC Score}
The Forecast AC Score occupies a distinct position relative to both traditions.
Table~\ref{tab:comparison} summarizes the key design differences across the
metrics discussed above.
In the stability component, the AC Score shares the observation-independent
philosophy of Zsoter et al.\ \cite{zsoter2009jumpiness} and Griffiths et al.\
\cite{griffiths2019flipflop}, measuring revision consistency without requiring
knowledge of eventual outcomes. It departs from these predecessors in three
important respects.
First, the AC Score jointly evaluates accuracy and stability in a unified
framework rather than treating them as separate diagnostics to be combined post
hoc. The accuracy component directly addresses the limitation Ehret
\cite{ehret2010convergence} identified in purely observation-independent
metrics: absolute forecast quality is embedded in the score itself via the
energy score, rather than requiring an auxiliary measure.
Second, and most significantly, the AC Score is defined for the full predictive
distribution rather than for point forecasts or ensemble means treated as scalar
quantities. The stability component measures revision jumps via the squared
energy distance between successive marginal forecast distributions, capturing
instability in both the center and the spread of the predictive distribution
across revision origins. Two forecast sequences may be indistinguishable by any
point-forecast stability metric while exhibiting substantially different
uncertainty dynamics — for example, one system may maintain a consistent mean
trajectory while alternating between narrow and wide predictive intervals. The
AC Score penalizes such distributional instability; the Flip-Flop Index and
congruence measure do not. This generalization directly addresses the open
direction identified by Zsoter et al.\ \cite{zsoter2009jumpiness}, who called
for extending jumpiness measures to probabilistic forecasts. Compared to MQC
\cite{zanotti2025stability}, which partially addresses this gap using a finite
quantile representation, the AC Score operates on the full distribution and
inherits the theoretical properties of the energy distance as a proper metric on
the space of probability distributions.
Third, the horizon weighting scheme embedded in the weighted energy distance and
energy score allows the relative importance of forecast horizons to be
configured according to the decision problem at hand. While Ehret
\cite{ehret2010convergence} also incorporates lead-time weighting via an
exponential weighting exponent, that design is motivated by operational flood
management conventions. The AC Score's weighting is more general, accommodating
arbitrary horizon importance profiles suited to domains such as energy dispatch,
supply chain planning, or financial risk management.

\begin{table}[ht]
\centering
\caption{Comparison of forecast stability metrics.}
\label{tab:comparison}
\begin{tabular}{lcccc}
\hline
\textbf{Metric} & \textbf{Obs.-indep.} & \textbf{Probabilistic} & \textbf{Accuracy-informed} & \textbf{Horizon-aware} \\
\hline
Jumpiness Index \cite{zsoter2009jumpiness} & \checkmark & & & \\
Flip-Flop Index \cite{griffiths2019flipflop} & \checkmark & & & \\
Convergence Index \cite{ehret2010convergence} & & & \checkmark & \checkmark \\
Congruence \cite{pritularga2024} & \checkmark & & & \\
MQC \cite{zanotti2025stability} & \checkmark & partial & & \\
\textbf{AC Score (ours)} & & \checkmark & \checkmark & \checkmark \\
\hline
\end{tabular}
\end{table}

In addition to being an evaluation metric, the forecast AC score is
differentiable and thus can be directly used as the training objective in
gradient descent based optimizers. Section \ref{sec:experiments} demonstrates
this capability by using the AC score as the loss function for training a
differentiable SARI model via AdamW. 

\section{Proposed Framework}

\subsection{Rolling Forecast Ensemble}
Consider a multi-horizon forecasting scenario where the forecaster receives new
information at each timestamp $t$. The information available to the forecaster
at time $t$ is denoted as $\mathcal{Z}_t = (Z_1, Z_2, \cdots, Z_t)$. $Z_t$
represents the new information that arrives at $t$, which contains the newly
realized value of the target time series and any covariates that become known at
time $t$. The forecaster then generates $m$-step-ahead forecasts using all
relevant information available up to (and including) time $t$. The
$m$-step-ahead forecasts issued at \emph{origin time} $t$ are denoted as
$\hat{y}_{t+1|t}, \hat{y}_{t+2|t}, \cdots, \hat{y}_{t+m|t}$, where the
subscripts $t+j$ correspond to the forecast target time and $t$ corresponds to
the forecast origin. We assume the $m$-step-ahead forecasts are sampled from a
joint probability distribution $f_t(\hat{y}_{t+1|t}, \hat{y}_{t+2|t}, \cdots,
\hat{y}_{t+m|t})$.


We now consider the rolling forecasts over all forecast origins. Collecting
forecast distributions issued at each origin $t = 1,\dots, n$, we have
    \begin{equation}\label{eq:forecast block}
        \bF|\mathcal{Z} =
        \begin{bmatrix}
            & f_1(\hat{y}_{2|1}, \hat{y}_{3|1}, \cdots, \hat{y}_{m+1|1} | Z_1) \\
            & f_2(\hat{y}_{3|2}, \hat{y}_{4|2}, \cdots, \hat{y}_{m+2|2} | Z_2) \\
            & \vdots \\
            & f_n(\hat{y}_{n+1|n}, \hat{y}_{n+2|n}, \cdots, \hat{y}_{n+m|n} | Z_{n}) \\
        \end{bmatrix}
    \end{equation}
which represents the probability distribution of the entire rolling forecasts.

\begin{definition}[Forecast Ensemble]
    We repeatedly draw $k$ samples from the joint distribution
    $\bF|\mathcal{Z}$. Each sample has the form of a $n\times m$
    matrix:\begin{equation}\label{eq:forecast_sample_ensemble}
        S^{(j)} = \begin{bmatrix}
            \hat{y}_{2|1}^{(j)} & \hat{y}_{3|1}^{(j)} & \hat{y}_{4|1}^{(j)} & \cdots & \hat{y}_{m+1|1}^{(j)} \\
            \hat{y}_{3|2}^{(j)} & \hat{y}_{4|2}^{(j)} & & & \vdots \\
            \hat{y}_{4|3}^{(j)} & & \ddots & & \\
            \vdots & & & & \\
            \hat{y}_{n+1|n}^{(j)} & \cdots & & & \hat{y}_{n+m|n}^{(j)}
        \end{bmatrix}.
    \end{equation}
    We call the collection of $S^{(j)}, j=1,2,\cdots, k$, a forecast ensemble,
    which represents the distribution of the multi-horizon rolling forecast. We
    call each $S^{(j)}$ a multi horizon rolling forecast sample.
\end{definition}

\subsection{Forecast Stability}
Now we define the notion of forecast stability, which concerns how consistently
a forecasting system predicts the \emph{same} future event as the forecast
origin changes.

Qualitatively speaking, a forecasting method is said to be \emph{stable} if the
revisions between forecasts with the same target time are small as new
information arrives. Conversely, large updates between successive forecasts
indicate instability: the system repeatedly alters its expectation of the same
future target. Below we describe how to measure such variation in a natural way
so as to quantify forecast stability.

\begin{definition}[Vertical Stability]
The anti-diagonals in each sample $S^{(j)}$ of the forecast ensemble
\eqref{eq:forecast_sample_ensemble} represent forecasts issued at different
origins targeting the same timestamp, i.e., the entries
\begin{equation} \label{eq:antidiagonals}
\left\{ \hat{y}_{t|t-m}, \hat{y}_{t|t-m+1}, \cdots, \hat{y}_{t|t-1} \right\}
\end{equation}
are forecasts targeting the time $t$ issued at different origins $t-m, t-m+1,
\cdots, t-1$, respectively. Each anti-diagonal therefore collects the sequence
of forecast revisions for a particular target time. Vertical stability refers to
the volatility of the sequence \eqref{eq:antidiagonals}, measured by its
variance averaged over t. Figure \ref{fig:vertical_stability_concept} visualizes
the concept of vertical stability.

\begin{figure}[h]
\centering
\begin{tikzpicture}[
    node/.style={circle, draw, minimum size=1.1cm, inner sep=0pt},
    origin/.style={node, fill=teal!20, draw=teal!60},
    target/.style={node, fill=gray!15, draw=gray!50},
    forecast/.style={node, minimum size=1.0cm},
    arr/.style={->, >=stealth, thick}
]
\node[origin] (o1) at (0, 0)    {$y_1$};
\node[origin] (o2) at (2.5, 0)  {$y_2$};
\node[target] (oh) at (7.5, 0)  {$y_h$};
\node[forecast, fill=orange!20, draw=orange!60] (fh1) at (7.5, 3.8) {$\hat{y}_{h|1}$};
\node[forecast, fill=orange!20, draw=orange!60] (fh2) at (7.5, 2.2) {$\hat{y}_{h|2}$};
\draw[arr, color=teal!70]   (o1) to[out=30, in=210] (fh1);
\draw[arr, color=teal!70] (o2) to[out=40, in=220] (fh2);
\draw[dashed, gray] (oh) -- (fh1);
\draw[dashed, gray] (fh1) -- (fh2);
\node[below=0.25cm of o1,  font=\small] {origin 1};
\node[below=0.25cm of o2,  font=\small] {origin 2};
\node[below=0.25cm of oh,  font=\small] {target $h$};
\end{tikzpicture}
\caption{Vertical stability: forecasts $\hat{y}_{h|1}$ and $\hat{y}_{h|2}$ both target the same future time $h$ but are issued from different origins $1$ and $2$. An accurate and stable forecast model should produce $\hat{y}_{h|j}$ approaching $y_h$ while minimizing discrepancies between them.}
\label{fig:vertical_stability_concept}
\end{figure}

In applications such as supply chain demand forecasting, having demand forecasts
of the same target day issued at different origins show great variability can be
problematic due to the decision making costs. Certain decisions made based on
previous forecasts, e.g., ordering supplies from a supplier, incur real costs
and may not be reversible. Having greater volatility in forecasts targeting the
same timestamp can therefore make decision making difficult and reduce the
decision maker's confidence in the forecasts. In such situations, stability, in
addition to accuracy, is a desirable property for forecasts targeting the same
timestamp. This kind of stability was first raised in \cite{godahewa2025stability}
and named \emph{vertical stability}.
\end{definition}

\begin{definition}[Horizontal Stability]
Rows of the forecast ensemble \eqref{eq:forecast_sample_ensemble} correspond to
forecasts issued at a fixed origin time, i.e., all forecasts generated at time
$t$ with horizons ranging from 1 to $m$:
\[
\bF_{t,:} =
\begin{bmatrix} \hat{y}_{t+1|t}, \hat{y}_{t+2|t}, \cdots, \hat{y}_{t+m|t} \end{bmatrix}.
\]
This row therefore represents the \emph{forecast trajectory} or \emph{forecast
vector} produced at a single origin. Stability along this direction is called
\emph{horizontal stability} as proposed in \cite{godahewa2025stability}.
\end{definition}

\subsection{Measuring the Distance between Forecast
Distributions}\label{subsec:metric_definition}

Consider the space of all probability distributions. We may choose a metric
which evaluates the discrepancy between two distributions, including their
center, spread, and overall shape. 
\begin{example}[CRPS]
    Given two 1-dimensional random variables $X, Y$ with probability
    distributions $F$ and $G$ respectively, we can define the (squared) metric
    $d^2$ to be 
    \begin{equation}\label{eq:CRPS}
        d^2(F, G) = \E_{F,G}|X-Y| - \frac{1}{2} \E_{F}|X-X^\prime| - \frac{1}{2} \E_{G} |Y-Y^\prime|
    \end{equation}
    where $X^\prime$ is an independent and identical copy of $X$. Let $\delta_y$
    be the Dirac distribution supported at the true outcome value $y$. Then we
    may measure the accuracy of a forecast distribution by $d^2(\hat{Y},
    \delta_y)$, which is exactly the \emph{Continuous Ranked Probability Score
    (CRPS)} \cite{GneitingRaftery2007}. For 1-dimensional random variables,
    $CRPS$ can be expressed alternatively by
    \begin{equation}
        CRPS(F,y) = \int \left(F(x) - \mathbb{1}_{x>y}(y)\right)^2 dy.
    \end{equation}
\end{example}

\begin{example}[Energy Score and Distance]\label{example:energy_distance}
    In higher dimensional cases, we replace the absolute value in
    \eqref{eq:CRPS} by the $L^2$ norm, i.e.,
    \begin{equation*}
        D^2(F, G) = \E_{X \sim F, Y \sim G}\|X-Y\| - \frac{1}{2} \E_{X, X^\prime \sim F}\|X-X^\prime\| - \frac{1}{2} \E_{Y, Y^\prime \sim G} \|Y-Y^\prime\|
    \end{equation*}
    where $\|(x_1, \cdots, x_n)\| = (\sum x_i^2)^\frac{1}{2}$ is the $L^2$ norm.
    Then we recover the squared \emph{Energy Distance}, see
    \cite{Szekely2003Estats} and \cite{rizzo2016energy}. Suppose the random
    variable $Y\sim G$ is realized with realized value $y$. One may view the
    realization as a Dirac measure supported at $y$. Substituting this it into
    $G$, we recover the \emph{Energy Score} \cite{GneitingRaftery2007}
    \begin{equation}
        ES(F, y) = \E_{X \sim F} \|X-y\| - \frac{1}{2} \E_{X, X^\prime \sim F} \|X - X^\prime\|.
    \end{equation}
    Note that the energy score reduces to CRPS in the 1-dimensional case.
\end{example}

Choosing a suitable metric allows us to study how the sequence of forecast
revisions evolves as the forecast origin changes. To be precise, we fix a target
time $t$ and consider two sequences of forecast paths $\hat{y}_{t|s}$ and
$\hat{y}^\prime_{t|s}$, both aimed at predicting the same realization at time
$t$. Let $y$ denote the true outcome of the target.
\begin{figure}[ht]
    \centering
    \begin{tikzpicture}[scale=2]

\def\norigins{4}          
\def\radiusStep{0.5}      

\def\stablecolor{blue!60}       
\def\volatilecolor{red!60}      
\def\circlecolor{gray!30}       

\foreach \i in {1,...,\norigins} {
    \draw[\circlecolor, dashed, thick] (0,0) circle (\i*\radiusStep);
}

\coordinate (g1) at ({\radiusStep*1*cos(80)},{\radiusStep*1*sin(80)});
\coordinate (g2) at ({\radiusStep*2*cos(82)},{\radiusStep*2*sin(82)});
\coordinate (g3) at ({\radiusStep*3*cos(78)},{\radiusStep*3*sin(78)});
\coordinate (g4) at ({\radiusStep*4*cos(81)},{\radiusStep*4*sin(81)});

\draw[\stablecolor, thick, -] (g1) -- (g2) -- (g3) -- (g4);
\foreach \i/\coord/\label in {1/g1/$\hat{y}_{5|4}$, 2/g2/$\hat{y}_{5|3}$, 3/g3/$\hat{y}_{5|2}$, 4/g4/$\hat{y}_{5|1}$} {
    \node[\stablecolor, fill=\stablecolor, circle, inner sep=1.5pt,
          label={[font=\small, above right,xshift=2pt,yshift=2pt]\label}] at (\coord) {};
}

\coordinate (f1) at ({-\radiusStep*1*cos(40)},{\radiusStep*1*sin(40)});
\coordinate (f2) at ({-\radiusStep*2*cos(70)},{\radiusStep*2*sin(70)});
\coordinate (f3) at ({-\radiusStep*3*cos(30)},{\radiusStep*3*sin(30)});
\coordinate (f4) at ({-\radiusStep*4*cos(75)},{\radiusStep*4*sin(75)});

\draw[\volatilecolor, thick, -] (f1) -- (f2) -- (f3) -- (f4);
\foreach \i/\coord/\label in {1/f1/$\hat{y}^\prime_{5|4}$, 2/f2/$\hat{y}^\prime_{5|3}$, 3/f3/$\hat{y}^\prime_{5|2}$, 4/f4/$\hat{y}^\prime_{5|1}$} {
    \node[\volatilecolor, fill=\volatilecolor, circle, inner sep=1.5pt,
          label={[font=\small, below left,xshift=-2pt,yshift=-2pt]\label}] at (\coord) {};
}

\node[black, fill=black, circle, inner sep=2pt,
      label={[font=\small, left]$y$}] at (0,0) {};

\end{tikzpicture}
    \caption{Illustration of two forecast sequences.}
    \label{fig:forecast_paths_1}
\end{figure}

Suppose that both sequences have the same accuracy at every revision, in the
sense that 
\[ d\!\left(\hat{y}_{t|s},\, \delta_y\right) = d\!\left(\hat{y}^\prime_{t|s},\,
\delta_y\right) \qquad \text{for all valid origins\;} s. 
\] 
As shown in Figure~\ref{fig:forecast_paths_1}, even with equal accuracy, the
\emph{paths} traced by the two sequences may differ in volatility. We prefer the
sequence $\{\hat{y}_{t|s}\}$ because its path is less volatile. This motivates
separating the notions of \emph{accuracy} (closeness to the truth) and
\emph{stability} (smoothness of the forecast path).

\begin{definition}[Accuracy score defined on joint distributions]
Our score consists of two components, an accuracy component
$\Acc(F|\mathcal{Z})$, which measures the accuracy of the multi-horizon
forecasts, and a stability component $\Stb(F|\mathcal{Z})$, which measures the
smoothness between successive forecasts.

The accuracy score is defined via the energy score. Let 
\[
    Y_{ \text{true}}(t) = (y_{t+1}, \cdots, y_{t+m})
\]
denote the true values of the next $m$ target values. Let 
\[
    \hat{Y}_{\text{pred}}(t) = (\hat{y}_{t+1|t}, \cdots, \hat{y}_{t+m|t})
\]
be the multivariate random variable representing the next $m$ step forecasts.
Then the weighted energy score of the multi-step forecast distribution $f_t$
given true outcomes $y_{\text{true}}(t)$ is equal to
\begin{equation}\label{eq:energy_score_fixed_origin}
    \ES(f_t, Y_{\text{true}}(t)) = \E\|\hat{Y}_{\text{pred}}(t) - Y_{\text{true}}(t)\|_w - \frac{1}{2} \E\|\hat{Y}_{\text{pred}}(t) - \hat{Y}^\prime_{\text{pred}}(t)\|_w,
\end{equation}
where $\hat{Y}_{\text{pred}}^\prime(t)$ is an independent copy of
$\hat{Y}_{\text{pred}}(t)$, i.e., $\hat{Y}_{\text{pred}}^\prime(t)$ has the same
distribution as but is independent of $\hat{Y}_{\text{pred}}(t)$. Here
$\|\cdot\|_w$ is a weighted version of Euclidean distance, i.e.,
\begin{equation}\label{eq: weighted_euclidean_norm}
    \|X-Y\|_w = \left(\sum_i w_i|X_i - Y_i|^2 \right)^\frac{1}{2},
\end{equation}
with $w_i \geq 0$. The purpose of the weight $w$ is to assign different levels
of importance to optimizing the accuracy of short-range forecasts versus
long-range forecasts.
The \emph{accuracy score} is defined by 
\begin{equation}\label{eq:energy_score_time_average}
    \Acc(F|\mathcal{Z}) = \frac{1}{n} \sum_{t=1}^n \ES(f_t, y_{\text{true}}(t)).
\end{equation}
It measures the average accuracy across all forecast origin times.
\end{definition}

\begin{remark}[Generalization to Non-Euclidean Target Spaces]
The accuracy score as defined above uses the weighted Euclidean norm
$\|\cdot\|_w$ in \eqref{eq: weighted_euclidean_norm} to measure distances
between forecast distributions and ground truth. While this choice is natural
when the target values lie in $\R^m$, many practical forecasting problems
involve targets that are constrained to a submanifold of Euclidean space. For
example, the Circular Flip-Flop Index of Griffiths et al.\
\cite{griffiths2021circular} was designed to evaluate wind direction forecasts,
which lie on the unit circle. In such cases, the Euclidean distance may fail to
respect the geometry of the target space. For example, it would treat the
angular difference between $359^\circ$ and $1^\circ$ as large, when the true
geodesic distance on the circle is small.

Our framework extends naturally to this setting. In fact, let $\mathcal{M}
\subset \R^d$ be a Riemannian submanifold equipped with distance
$d_{\mathcal{M}}$. One may replace $\|\cdot\|_w$ in
\eqref{eq:energy_score_fixed_origin} and
\eqref{eq:energy_distance_two_distributions} with the corresponding weighted
intrinsic distance
\[
    \|X - Y\|_{\mathcal{M}, w} = \left(\sum_i w_i\, d_{\mathcal{M}}(X_i, Y_i)^2\right)^{1/2},
\]
provided $d_{\mathcal{M}}$ is a negative definite kernel. The negative definite
kernel condition is required for the energy distance to remain a valid metric on
the space of probability distributions, see \cite{rizzo2016energy}. The unit
circle $S^1$ with arc-length distance, and more generally the unit sphere
$S^{n-1}$, satisfy this condition. Wind direction forecasting is a natural
application: target values lie on $S^1$, and the Circular Flip-Flop Index of
Griffiths et al.\ \cite{griffiths2021circular} addresses precisely this setting
for point forecasts. Our framework provides the analogous generalization for
full predictive distributions.
\end{remark}

We now proceed to define the stability measurement of our forecast ensemble. To
compare forecasts issued at successive origins $t$ and $t+1$, we need to align
them to a common set of target times. The forecast from origin $t$ covers
targets $t+1, \ldots, t+m$, while the forecast from origin $t+1$ covers $t+2,
\ldots, t+m+1$. Their overlap is the window $t+2, \ldots, t+m$. We introduce the
following notation to make this alignment explicit.
 
\begin{definition}[Window-restricted forecast]
    Let $\hat{Y}_{\text{pred}}(t;\ t+a, t+b)$ denote the joint forecast
    distribution issued at origin $t$ and targeting the contiguous window of
    times $\{t+a, t+a+1, \ldots, t+b\}$, where $1 \leq a \leq b \leq m$. That
    is,
    \[
        \hat{Y}_{\text{pred}}(t;\ t+a, t+b) = (\hat{y}_{t+a|t},\ \hat{y}_{t+a+1|t},\ \ldots,\ \hat{y}_{t+b|t}).
    \]
    The full $m$-step forecast issued at origin $t$ is then
    $\hat{Y}_{\text{pred}}(t;\ t+1, t+m)$.
\end{definition}
 
With this notation, the overlapping window shared by forecasts from origins $t$
and $t+1$ is simply $\hat{Y}_{\text{pred}}(t;\ t+2, t+m)$ and
$\hat{Y}_{\text{pred}}(t+1;\ t+2, t+m)$ respectively — both targeting the same
set of future times $t+2, \ldots, t+m$, and therefore directly comparable.
 
\begin{definition}[Stability score defined on joint distribution given realization]
    The \emph{stability score} measures the smoothness of the forecast revision
    path, i.e., how much the forecasts targeting the same timestamp change
    between successive origins. The forecast issued at origin $t$ restricted to
    the overlap window is $\hat{Y}_{\text{pred}}(t;\ t+2, t+m)$, and likewise
    $\hat{Y}_{\text{pred}}(t+1;\ t+2, t+m)$ for origin $t+1$. The difference
    between these two forecast distributions can be measured by the squared
    weighted energy distance
    \begin{equation}\label{eq:energy_distance_two_distributions}
        \begin{aligned}
            & D^2\bigl(\hat{Y}_{\text{pred}}(t;\ t+2,t+m),\, \hat{Y}_{\text{pred}}(t+1;\ t+2,t+m)\bigr) \\
            &= \E\bigl\|\hat{Y}_{\text{pred}}(t;\ t+2,t+m) - \hat{Y}_{\text{pred}}(t+1;\ t+2,t+m)\bigr\|_{w} \\
            &\quad - \frac{1}{2}\,\E\bigl\|\hat{Y}_{\text{pred}}(t;\ t+2,t+m) - \hat{Y}^{\prime}_{\text{pred}}(t;\ t+2,t+m)\bigr\|_{w} \\
            &\quad - \frac{1}{2}\,\E\bigl\|\hat{Y}_{\text{pred}}(t+1;\ t+2,t+m) - \hat{Y}^{\prime}_{\text{pred}}(t+1;\ t+2,t+m)\bigr\|_{w},
        \end{aligned}
    \end{equation}
    where $\hat{Y}^{\prime}_{\text{pred}}$ denotes an independent copy of
    $\hat{Y}_{\text{pred}}$. We average over all pairs of successive forecast
    origins to get an overall stability score:
    \begin{equation}\label{eq:stability_score_time_average}
        \Stb(F|\mathcal{Z}) = \frac{1}{n-1}\sum_{t=1}^{n-1} D^2\bigl(\hat{Y}_{\text{pred}}(t;\ t+2,t+m),\, \hat{Y}_{\text{pred}}(t+1;\ t+2,t+m)\bigr).
    \end{equation}
    This characterizes the expected stability score as an average over all
    forecast samples generated for a single realization.
\end{definition}
 
\begin{definition}[Forecast Accuracy and Coherence (AC) Score]
    The \emph{forecast accuracy and coherence score}, or \emph{AC score} for
    short, is defined as a weighted combination of accuracy and stability
    scores:
    \begin{equation}\label{eq:total_score}
        S_{\AC; \lambda}(F|\mathcal{Z}) = \Acc(F|\mathcal{Z}) + \lambda \Stb(F|\mathcal{Z})
    \end{equation}
    for some non-negative weight $\lambda$. Typically we suppress the dependence
    of $\lambda$ in our notation, using $S_\AC$ for brevity.
\end{definition}
 
\begin{remark}
The metric evaluates forecast quality based solely on the resulting
distributions, without assuming how those forecasts are produced. This can be
applied to diverse types of multi-horizon forecasting models, for example
\begin{itemize}
    \item \textit{Iterative forecasting:} Multi-step forecasts generated by
    repeatedly applying one-step-ahead models, common in classical time series
    methods and transformer-based architectures. The ARIMA model used the
    experiments of this paper is an example of iterative forecasting. See
    \cite{seabold2010statsmodels}.
    \item \textit{Direct multi-horizon forecasting:} Models trained to produce
    all forecast horizons simultaneously, which is commonly used in modern
    neural architectures. Notable examples include Informer
    \cite{zhou2021informer}, which uses a generative style decoder to predict
    long output sequences in one step, and N-BEATS \cite{oreshkin2019n}, a
    purely feed-forward architecture with interpretable trend and seasonality
    decomposition.
    \item \textit{Skeleton-based forecasting:} Generating forecasts at sparse
    key horizons (e.g., $h \in \{24, 48, 72\}$ for hourly data) and using
    interpolation, extrapolation, or learned smoothing to obtain intermediate
    forecasts. 
    \item \textit{Attention-based architectures:} Transformer models like
    Temporal Fusion Transformer \cite{lim2021temporal} that use self-attention
    mechanisms to capture temporal dependencies and generate multi-horizon
    forecasts.
\end{itemize}
\end{remark}
\subsection{Empirical Approximation}
The scores defined in Section \ref{subsec:metric_definition} can be computed
empirically with samples. Suppose the forecaster makes forecasts on a rolling
basis. At each time $t$, the forecaster utilizes the most up-to-date information
at $t$, namely $Z_t$, to generate $k$ sample paths $(\hat{y}_{t+1|t}^{(i)},
\cdots, \hat{y}_{t+m|t}^{(i)})$, $1 \leq i \leq k$, each of which forecasts the
next $m$ step target values. When the rolling forecasting procedure finishes, we
have $k$ forecast sample ensembles of the form
\eqref{eq:forecast_sample_ensemble}. We illustrate how to compute the accuracy
score \eqref{eq:energy_score_time_average} and the stability score
\eqref{eq:stability_score_time_average} empirically based on the sample
ensembles. The energy score \eqref{eq:energy_score_fixed_origin} can be computed
empirically by 
\begin{equation}\label{eq:energy_score_empirical}
    \begin{aligned}
        \frac{1}{k} \sum_{i=1}^{k} \left( \sum_{j=1}^{m} w_j |\hat{y}^{(i)}_{t+j|t} -  y_{t+j}|^2 \right)^{\frac{1}{2}} - \frac{1}{k(k-1)} \sum_{1 \leq i_1 < i_2 \leq k} \left( \sum_{j=1}^{m} w_j |\hat{y}^{(i_1)}_{t+j|t} - \hat{y}^{(i_2)}_{t+j|t}|^2 \right)^{\frac{1}{2}}.
    \end{aligned}
\end{equation}

Similarly, the energy distance \eqref{eq:energy_distance_two_distributions}
between successive forecasts can be computed empirically by
\begin{equation}\label{eq:energy_distance_empirical}
    \begin{aligned}
        & \frac{1}{k} \sum_{i=1}^k \left(\sum_{j=2}^m w_j |\hat{y}_{t+j|t}^{(i)} - \hat{y}_{t+j|t+1}^{(i)}|^2 \right)^\frac{1}{2} \\
        & - \frac{1}{k(k-1)} \sum_{1 \leq i_1<i_2\leq k} \left(\sum_{j=2}^m w_j |\hat{y}_{t+j|t}^{(i_1)} - \hat{y}_{t+j|t}^{(i_2)}|^2 \right)^\frac{1}{2} \\
        & - \frac{1}{k(k-1)} \sum_{1 \leq i_1<i_2\leq k} \left(\sum_{j=2}^m w_j |\hat{y}_{t+j|t+1}^{(i_1)} - \hat{y}_{t+j|t+1}^{(i_2)}|^2 \right)^\frac{1}{2}.
    \end{aligned}
\end{equation}

Then, the accuracy score \eqref{eq:energy_score_time_average} and the stability
score \eqref{eq:stability_score_time_average} are computed by averaging
\eqref{eq:energy_score_empirical} and \eqref{eq:energy_distance_empirical}\ over
times $t$, respectively. 

\begin{remark}[Point Forecast as a Special Case]
    The forecast AC score can be extended to evaluating point forecasts. In
    fact, the point estimate $\hat{y}$ may be viewed as a Dirac distribution
    supported at $\hat{y}$. Thus the empirical formula of accuracy score
    \eqref{eq:energy_score_empirical} becomes 
    \begin{equation}
        Acc(F|Z) = \frac{1}{n} \sum_{t=1}^n \left(\sum_{j=1}^{m} w_j |\hat{y}_{t+j|t} -  y_{t+j}|^2 \right)^\frac{1}{2}
    \end{equation}
    and \eqref{eq:energy_distance_empirical} becomes
    \begin{equation}
        Stb(F|Z) = \frac{1}{n-1} \sum_{t=1}^{n-1} \left(\sum_{j=2}^m w_j |\hat{y}_{t+j|t} - \hat{y}_{t+j|t+1}|^2 \right)^\frac{1}{2}.
    \end{equation}
    
\end{remark}

\subsection{Expected AC Score Given a Data Generation Process}
So far, we have defined the accuracy and stability score where all samples of
forecasts are conditioned on the same realization. This applies to use cases
where the experiment that generated the realized time series is done only once,
e.g. sales numbers of a specific retail shop. In some situations, there are
multiple observed test subjects, each yielding their own realized time series,
all subject to the same underlying mechanism. Consider the following example
where someone working for Walmart has access to the sales data from Walmart
stores across the nation and their goal is to choose a single forecast model
that achieves the best overall performance for all stores. One might adopt the
simplifying assumption that the sales time series from all stores are generated
by the same data generation process with noise. This motivates extending the AC
score conditioned on a realization to the following. 
\begin{definition}[Forecast AC score given data generating process]
    Suppose the underlying data generation process is known. We define the
    expected stability score as
    \begin{equation}\label{eq:stability_score_averaged_over_realizations}
        S_\AC(F) = \E_{\mathcal{Z}}[S_\AC(F|\mathcal{Z})].
    \end{equation}
The expectation is taken over all possible realizations based on their
likelihood under the data generation process. This expected score characterizes
how well one expects the forecast model to work on a random test subject which
follows the same data generation process as the time series which the model was
trained on.
\end{definition}

\subsection{The Choice of Weights}\label{sec:time_discounting}
In this subsection, we discuss the potential choice of the weights $w_j$ in
\eqref{eq: weighted_euclidean_norm}. Not all forecast updates matter equally.
Early forecasts are highly uncertain and usually have limited operational
impact, while updates made close to a decision point directly affect actions
such as ordering, scheduling, or resource allocation. Therefore, both accuracy
and stability should be weighted more heavily near a decision point, where
errors and unnecessary revisions are most costly. Introducing time-dependent
weights allows the score to reflect these practical priorities: emphasizing
accuracy near but ahead of decision points, penalizing last-minute volatility,
and aligning the evaluation with the timing of real decisions.

By allowing accuracy and stability to use separate weighting functions, the
framework can be adapted to different operational contexts in which the value of
precise or stable forecasts changes over time. 

For a fixed target time $T$ and with the earliest origin denoted as $i_{\min}$,
let the decision time $\tau$ be a time between the earliest origin and the
target time, i.e., $i_{\min} \leq \tau < T$. Then, it is convenient to define
the distance to the decision time from the origin as $u=|i-\tau|$ and to define
a corresponding weighting function $w(u)$. Similarly, define the horizon $h$ as
the distance from origin to target time, $h=T-i$, and the corresponding
weighting function $w(h)$. Depending upon the operational context, one
representation may be more natural than the other, i.e., a representation based
on distance from the origin to the decision time versus a representation based
on distance from the origin to the target time. Below, we will develop the case
for $w(h)$; that for $w(u)$ is analogous.

To ensure a meaningful and interpretable evaluation, the weighting function
should satisfy a few basic properties:
\begin{enumerate}
    \item Non-negativity: $w(h) \ge 0$.
    \item Normalization: $\sum_{h} w(h)=1$. This ensures compatibility across
          horizons and across datasets.
    \item Monotonicity toward the target time (optional): weights increase as
          we approach the decision point, $w(h_1) \le w(h_2)$ whenever
          $h_1 > h_2$. It is assumed that later forecasts are more consequential.
          However, the framework allows for alternative shapes (e.g., U-shaped)
          depending upon domain needs.
\end{enumerate}
The weighting function is a hyperparameter; one is free to choose one that fits
a use case best. Below we summarize several widely used functional forms:
\begin{enumerate}
  \item Exponential: $w(h) = e^{-\alpha h}$
  \item Linear: $w(h) = a + b h, b<0$
  \item Hyperbolic: $w(h) = \frac{1}{1+\beta h}$
  \item Inverse-variance: $w(h) = \frac{1}{\mathrm{Var}(f_{h})}$
  \item Piecewise: weights change at predefined horizons (e.g., operational cycles)
\end{enumerate}
The functional forms listed above specify the relative shape of the weights but
do not enforce that they sum to one. In practice, the weights should be
normalized.

Accuracy and stability capture different aspects of forecast quality. Accuracy
measures closeness to the true target, while stability measures temporal
consistency across origins. Their operational relevance may differ: late
revisions can have high costs, affecting stability, while accuracy may be more
relevant across all horizons. The proposed framework allows the use of separate
weighting functions, $w(h)$ for accuracy and $w'(h)$ for stability, to reflect
these differences. In applications where the same importance applies, the same
function can be used for both.

\section{Experiments}\label{sec:experiments}

In this section, we extend the theoretical framework to practical application.
In machine learning, models are trained by optimizing a selected loss function
that quantifies the error or likelihood of their forecasts. During training on
historical data, the model learns optimal parameters by minimizing the error
loss (or maximizing the likelihood function). The choice of loss function is
critical because it directly shapes what patterns the model learns to capture
and prioritize. Common metric functions include root mean squared error (RMSE),
mean absolute error (MAE), mean absolute percentage error (MAPE), log
likelihood, and the continuous ranked probability score (CRPS) for probabilistic
forecasts.

Most implemented machine learning forecasting models employ one of these
standard metrics as their training objective. However, a significant limitation
of current practice is that loss evaluation typically focuses on one-step-ahead
forecast accuracy. This limitation is suboptimal in certain situations, as
optimizing for one step or a fixed horizon often results in a model that
performs well at the specific horizon but deteriorates at other horizons. In
fact, many practical applications require evaluating forecast quality across the
entire prediction window rather than at isolated time points. For example,
energy system operators may require reliable load forecasts from 1 hour ahead up
to 7 days ahead simultaneously in order to make operational decisions. A loss
function that treats each horizon independently can potentially lead to
suboptimal real-world performance despite strong performance on standard
benchmarks.

Another dimension of forecast quality that the traditional forecast evaluation
metrics fail to address is stability. Consider a supply chain planner who must
order inventory months in advance to meet anticipated demand. Suppose the
forecast made in January predicted June demand to be 10,000 units, while an
updated forecast made in February for that same June period might suddenly jump
to 14,000 units, then drop to 9,500 units in the March update. Such volatility
creates difficulty for decision making and planning. Procurement decisions are
often irreversible, as placed orders may not be canceled easily. Unstable
forecasts force planners into a difficult position: committing to large
purchases based on high forecasts risks excessive inventory holding costs and
potential obsolescence, while conservative ordering based on uncertain forecasts
risks stockouts and lost sales. As a result, large volatility can reduce the
confidence of the decision maker in the forecasts. Thus, in such situations
where commitments are made based on long-horizon forecasts, stability can be a
highly desirable property to have in addition to accuracy. However, in most
machine learning implementations, the objective function does not capture any
notion of stability. We note that \cite{vanbelle2023stability} has proposed an
extension to the N-BEATS deep learning architecture, which optimizes forecasts
from both a traditional forecast accuracy perspective as well as a forecast
stability perspective.

Our forecast AC score can address these limitations of traditional training
objectives by encouraging the model to optimize for both the accuracy and
stability of long horizon forecasts. When compared to models trained with
traditional metrics, models trained using our multi-horizon stability aware
score have the following advantages:
\begin{itemize}
    \item Accounting for accuracy over all horizons, instead of focusing on
    one-step-ahead forecasts.
    \item Accounting for vertical stability. The forecasts targeting the same
    timestamp are less volatile.
    \item Forecasters can easily adjust the horizon awareness of the score by
    assigning weights that penalize inaccuracy and instability at different
    horizons. In the extreme case where all but the one-step-ahead weights are
    set to zero, the metric is reduced to optimizing one-step-ahead forecasts.
    \item Forecasters can adjust the emphasis on stability via the stability
    parameter $\lambda$. 
\end{itemize}

In this section, we show how to set up a seasonal auto-regressive integrated
model with our metric as the objective function. The trained model is compared
with its counterpart trained using traditional objectives. Out-of-sample tests
are performed, in which we see that models trained using our metric achieve
better accuracy and stability over long horizons.

\subsection{Dataset}
We evaluate our method on the M4 Hourly dataset \cite{makridakis2018m4}, a
widely-used forecasting benchmark consisting of 414 hourly time series from
diverse domains. This dataset was introduced in the M4 forecasting competition
and has become a standard benchmark dataset for evaluating forecasting methods.

\subsection{Model Explanation}
We employ the seasonal autoregressive integrated moving average model (SARIMA).
SARIMA captures both non-seasonal and seasonal patterns in the data. Most of the
time series in the M4 dataset exhibit seasonal behavior. Thus, SARIMA models
serve as suitable baseline models for demonstrating the performance of our
metric. 

A SARIMA model is specified by seven hyperparameters:
SARIMA($p$,$d$,$q$)×($P$,$D$,$Q$,$s$) where:
\begin{itemize}
    \item $p$ is the order of autoregression, which represents how many past values influence the current value
    \item $d$ is the degree of differencing required to make the time series stationary
    \item $q$ is the order of the moving average, which represents the influence of past forecast errors
    \item $P$ is the seasonal autoregressive order
    \item $D$ is the seasonal differencing order
    \item $Q$ is the seasonal moving average order
    \item $s$ is the period of the season
\end{itemize}

For computational simplicity, we restrict the moving average order $q$ and $Q$
to be 0. In other words, our model takes the form of
SARIMA($p$,$d$,0)x($P$,$D$,0,$s$). We will refer to this model as SARI in the
paper. The mathematical representation of our model is given by
\begin{equation}\label{eq:SARIMA_math_representation}
    \Phi(L^s) \phi(L) (1 - L^s)^D (1 - L)^d y_t = \epsilon_t 
\end{equation}
where $L$ is the lag operator, which shifts the time series backward by one
step, i.e. $Ly_t = y_{t-1}$. The term $\phi(L)$ refers to the autoregressive
polynomial. That is, given autoregressive coefficients $\phi_1, \cdots, \phi_p$,

\[
    \phi(x) = 1 - \phi_1 x - \phi_2 x^2 - \cdots - \phi_p x^p.
\]
Similarly,
\[
    \Phi(x) = 1 - \Phi_1 x - \cdots \Phi_P x^P,
\]
For a fixed set of hyperparameters $(p, d, P, D, s)$, the model is determined by
the corresponding autoregressive and seasonal autoregressive coefficients
$\phi_j, \Phi_j$. This simple structure allows us to easily set up optimization
procedures to find the optimal coefficients under our metric.

\subsection{Training}
For each time series in the M4 Hourly dataset, we perform the following training
and evaluation procedure:

\textbf{1. Data Splitting}. We partition each series into training and test sets
using a 60-40 split ratio. Given a time series of length $T$, the training set
consists of the $[0, 0.6T]$ observations and the remaining observations go into
the test set for out-of-sample evaluation.

\textbf{2. Hyperparameter Selection}. We employ the auto-ARIMA algorithm from
the \texttt{statsforecast} library \cite{garza2022statsforecast} to
automatically determine the optimal SARI hyperparameters $(p, d, P, D, s)$. The
auto-ARIMA performs a stepwise search over the hyperparameter space. The search
space is constrained to $p \in \{0, 1, 2, 3\}$, $P \in \{0, 1\}$, and the
seasonal period is fixed at $s = 24$ to capture the daily cycle present in
hourly data. The Akaike Information Criterion (AIC) is used as the metric for
comparison, which accounts for both model fit and model complexity. This
automated hyperparameter selection procedure ensures that our models are tuned
to the specific pattern of each time series in the dataset.

\textbf{3. Traditional Model Fitting}. Using the hyperparameters found by
auto-ARIMA, we create a traditional SARI model using \texttt{statsmodels}
library \cite{seabold2010statsmodels}. The model is fitted on the training set
using maximum likelihood estimation (MLE) with standard one-step-ahead forecast
error minimization as the optimization objective. This serves as a benchmark for
comparison with the model optimized under our novel metric. The fitted MLE
coefficients — AR parameters, seasonal AR parameters, noise standard deviation,
and intercept — are saved and used to warm-start the differentiable model in the
next stage.

\textbf{4. Differentiable Model Fitting}. We start by creating a differentiable
SARI model with the same hyperparameters as the traditional model. The
autoregressive coefficients are treated as learnable parameters and warm-started
from the MLE solution obtained in the traditional fitting stage. Warm-starting
is essential because random initialization places the seasonal AR parameters far
from the MLE solution and causes the optimizer to spend a large number of epochs
in a flat loss region before converging. The MLE solution provides a good
initialization from which the AC score optimization makes fine-grained
adjustments.

Our innovation lies in using our multi-horizon stability-aware forecast AC score
as the objective for learning the autoregressive coefficients. In each training
epoch, we generate rolling forecasts on the training data. Specifically, the
rolling forecast starts from the timestamp whose index is equal to the minimum
required history length, which is in turn equal to the maximum lag of the SARI
model. We produce $H$-step-ahead forecasts for each valid forecast origin. The
multi-step forecasts are simulated via the equation
\eqref{eq:SARIMA_math_representation}. The forecast ensembles are stored in the
form of \eqref{eq:forecast_sample_ensemble}. The forecast origins are processed
in batches of size 64 for computational efficiency.

Next, we compute the loss by comparing the forecast ensemble against the true
outcomes according to our proposed metric using empirical formulas
\eqref{eq:energy_score_empirical} and \eqref{eq:energy_distance_empirical}. The
horizon aware weight $w$ was set to be linear 
\[
    w(j) = 1 - \frac{j}{h}
\]
and later normalized to sum up to 1. This weight encourages the model to pay
more attention to optimizing accuracy and stability of the short-range
forecasts, while maintaining a reasonable level of awareness of the quality of
long-range ones. The stability multiplier $\lambda$ is set to $0.5$.

Since the setup of the differentiable model as well as the entire forecast
generating process are implemented with PyTorch, the whole process is
differentiable. We backpropagate to compute the gradient of the loss function
$\nabla_{\phi, \Phi} S_{AC}(F|Z)$, where $\phi, \Phi$ refer to all of the
autoregressive coefficients. We employed the AdamW optimizer with an initial
learning rate of 0.001. A ReduceLROnPlateau scheduler was employed to reduce the
learning rate by a factor of 0.5 when the loss plateaus for 100 consecutive
epochs. Training runs for a minimum of 500 epochs and a maximum of 2000 epochs
and terminates early if the convergence criterion is met. We set the seed for
PyTorch and Numpy operations to 42 for reproducibility.

After the training is completed, we verify that the SARI model with the learned
coefficients is stationary by examining the roots of the autoregressive
polynomial 
\begin{equation}\label{eq:autoregressive_polynomial}
    \phi(L) \Phi(L^s).
\end{equation}
It is a well-known fact that a model is stationary if all roots of
\eqref{eq:autoregressive_polynomial} lie outside the unit circle $|z| = 1$.

\textbf{5. Out-of-sample Testing}. In order to validate the results on unseen
data, we perform out-of-sample tests. We use both the MLE trained model and the
model trained with the AC score to generate rolling probabilistic forecasts on
the test set (for each time series in the dataset). For each forecast origin in
the test set, we simulate $k = 30$ sample paths of length $H = 24$ steps. Both
models are evaluated from the same set of forecast origins, determined by the
minimum history offset, ensuring a fair comparison. The AC score, accuracy
score, and stability score are computed using the empirical formulas
\eqref{eq:energy_score_empirical} and \eqref{eq:energy_distance_empirical}. For
horizon-specific accuracy analysis, we additionally report the mean squared
error (MSE) of the ensemble mean forecast at each horizon, averaged across all
test series.

\subsection{Results}
\begin{figure}[htbp]
    \centering
    \begin{subfigure}[b]{0.4\textwidth}
        \centering
        \includegraphics[scale=0.3]{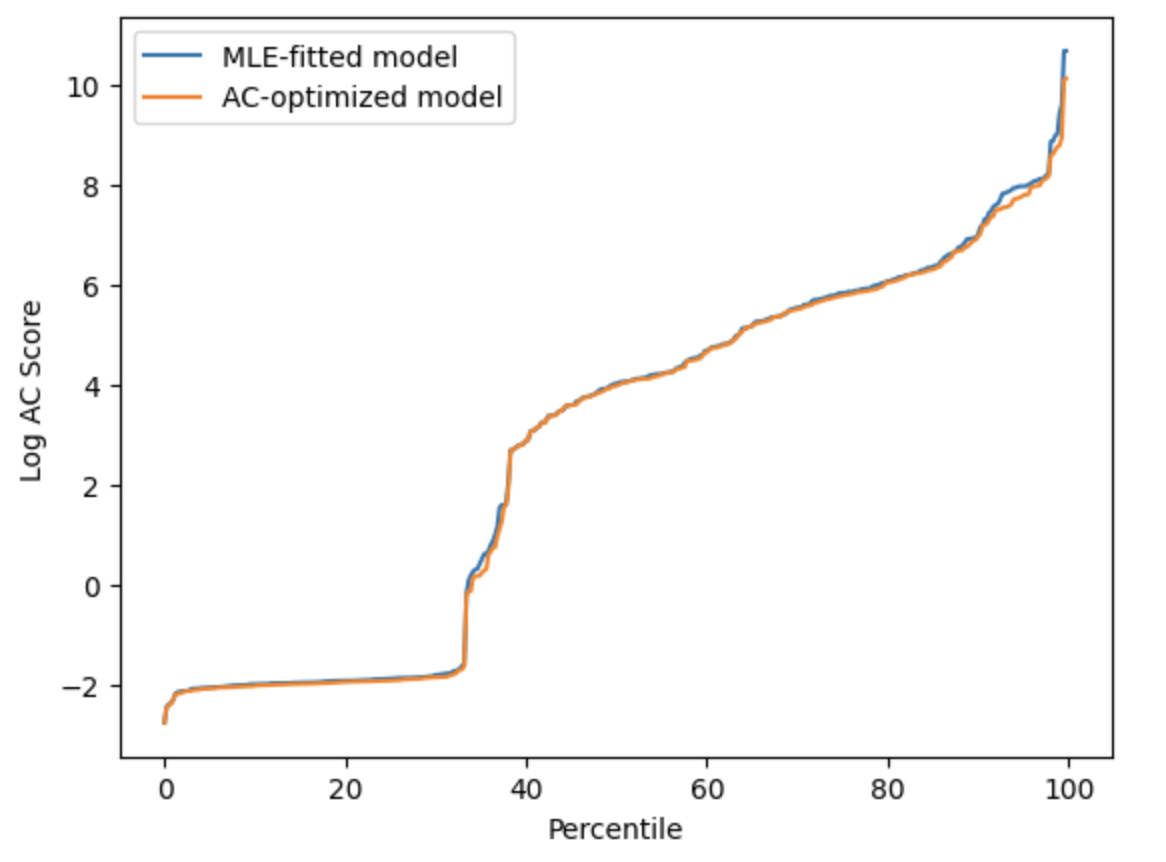}
        \caption{Log AC Score}
        \label{fig:log_AC_score}
    \end{subfigure}
    \hfill
    \begin{subfigure}[b]{0.4\textwidth}
        \centering
        \includegraphics[scale=0.3]{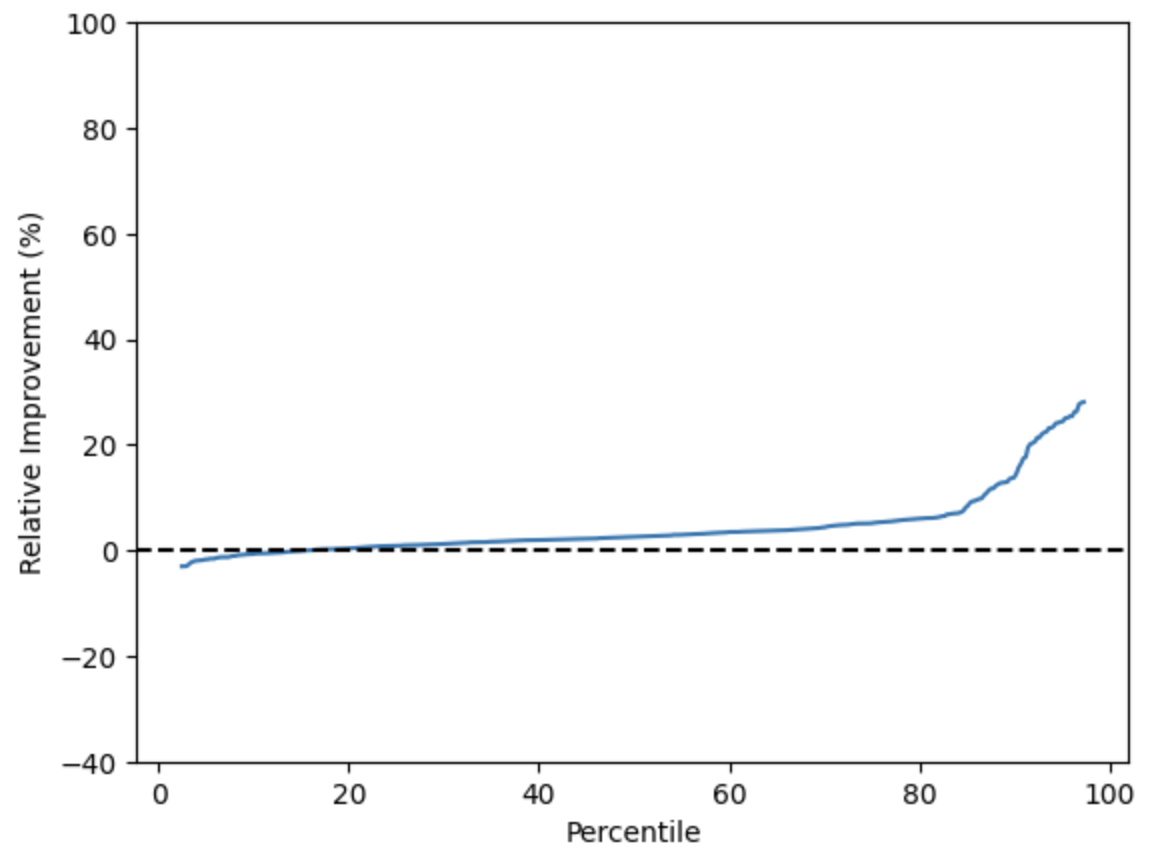}
        \caption{Relative Improvement of AC Score}
        \label{fig:rel_improv_AC_score}
    \end{subfigure}
    \vspace{0.5cm}
    \begin{subfigure}[b]{0.4\textwidth}
        \centering
        \includegraphics[scale=0.3]{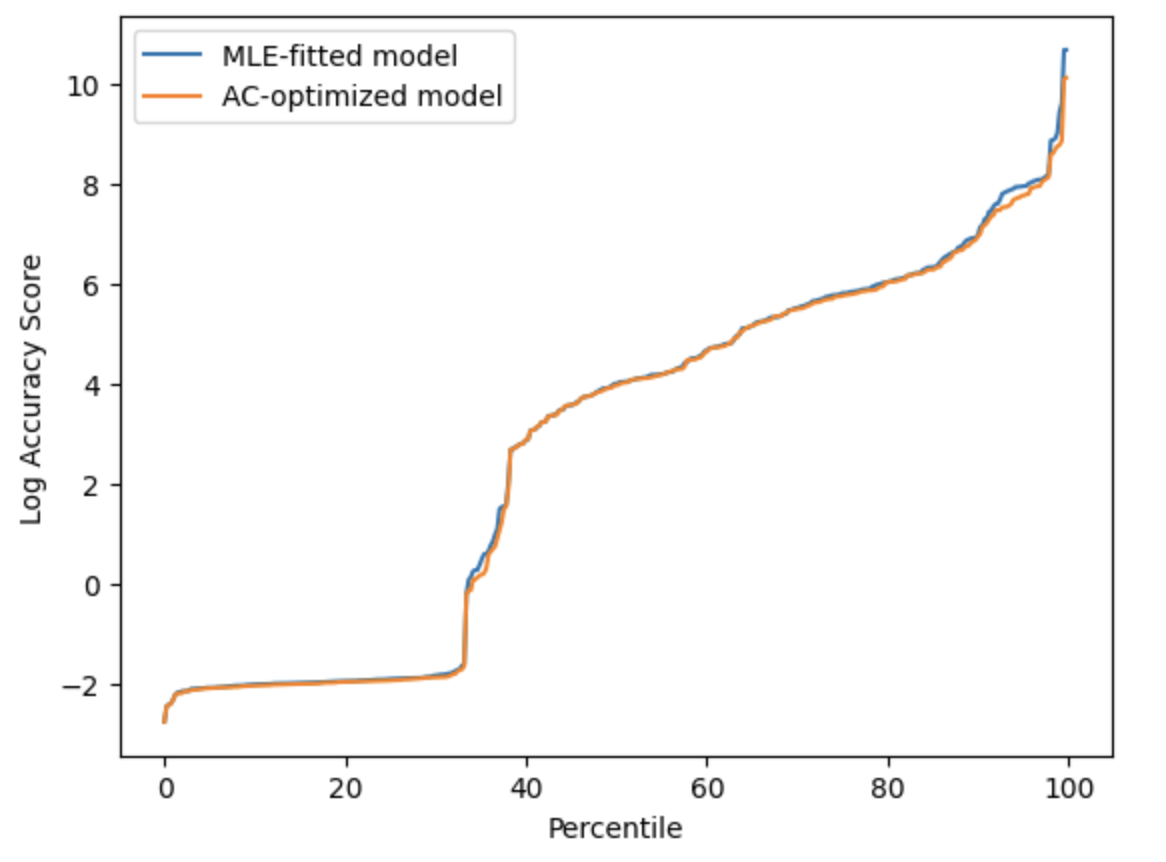}
        \caption{Log Accuracy Score}
        \label{fig:log_accuracy_score}
    \end{subfigure}
    \hfill
    \begin{subfigure}[b]{0.4\textwidth}
        \centering
        \includegraphics[scale=0.3]{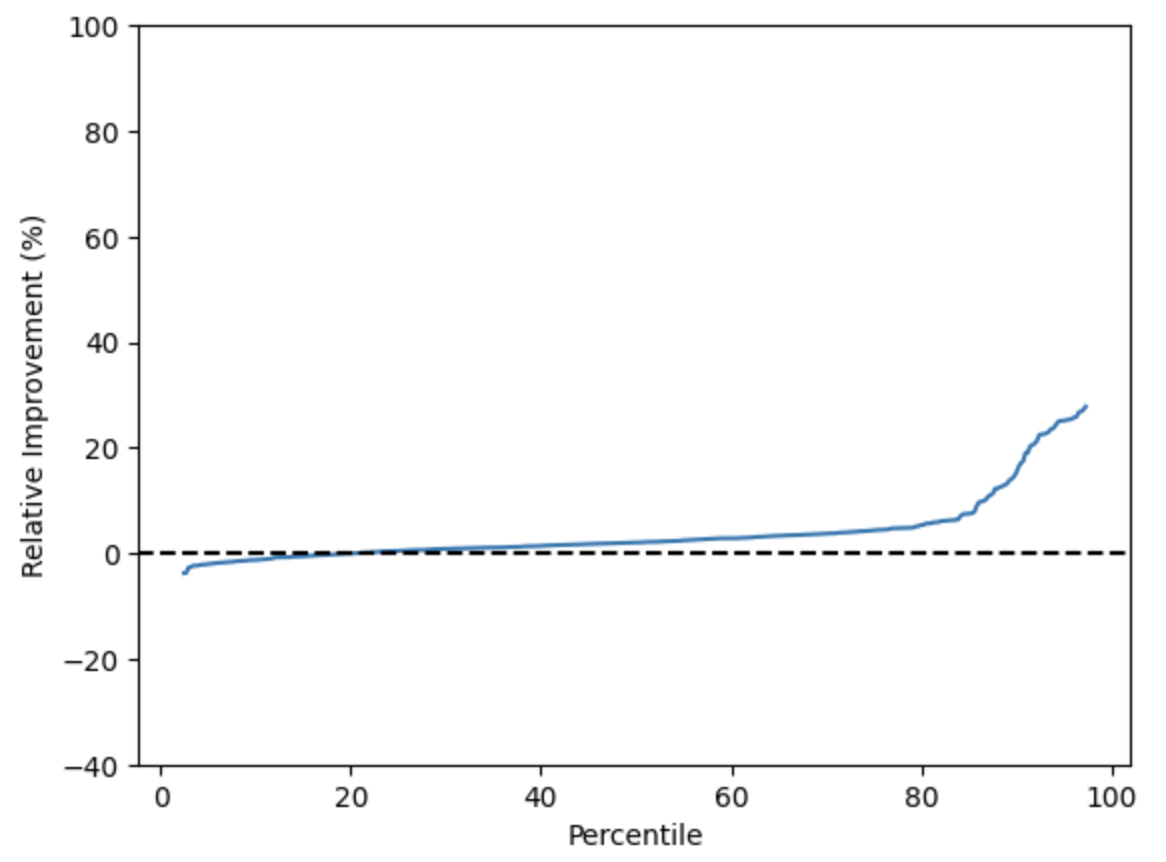}
        \caption{Relative Improvement of Accuracy Score}
        \label{fig:rel_improv_accuracy_score}
    \end{subfigure}
    \vspace{0.5cm}
    \begin{subfigure}[b]{0.4\textwidth}
        \centering
        \includegraphics[scale=0.3]{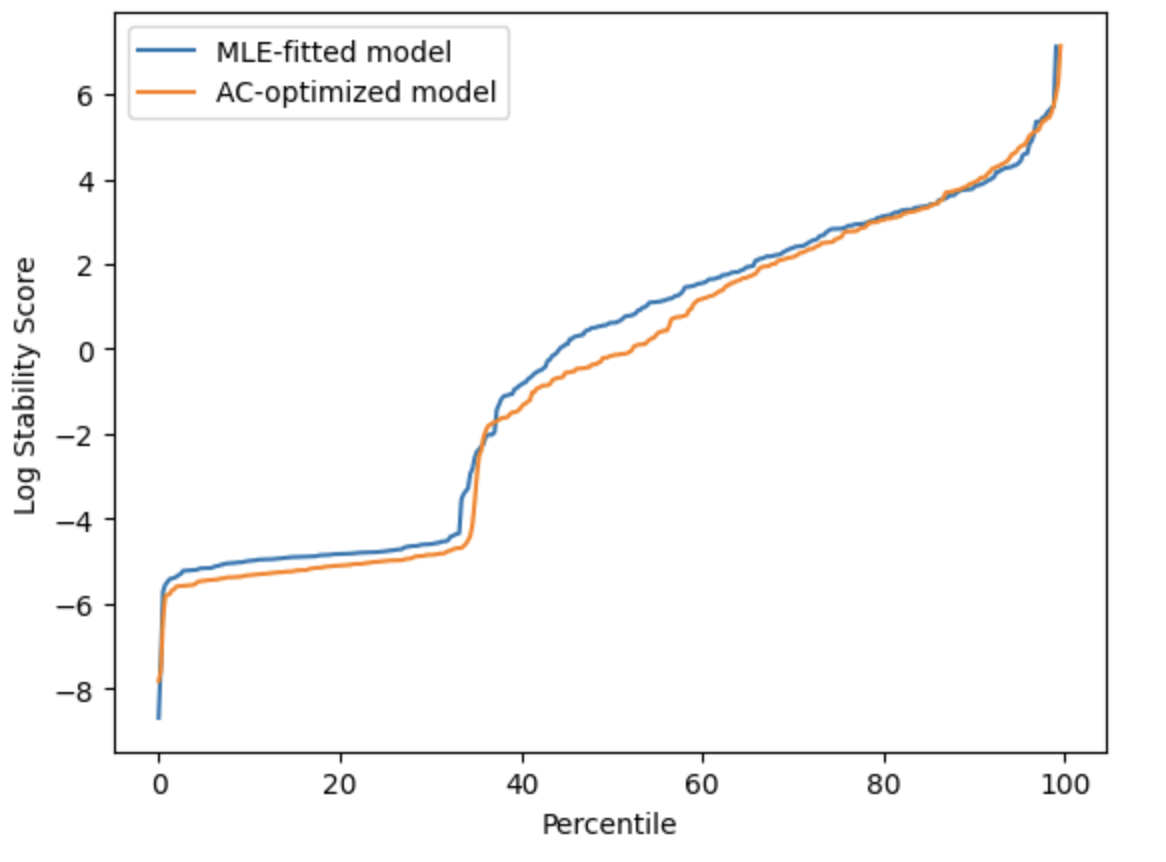}
        \caption{Log Stability Score}
        \label{fig:log_stability_score}
    \end{subfigure}
    \hfill
    \begin{subfigure}[b]{0.4\textwidth}
        \centering
        \includegraphics[scale=0.3]{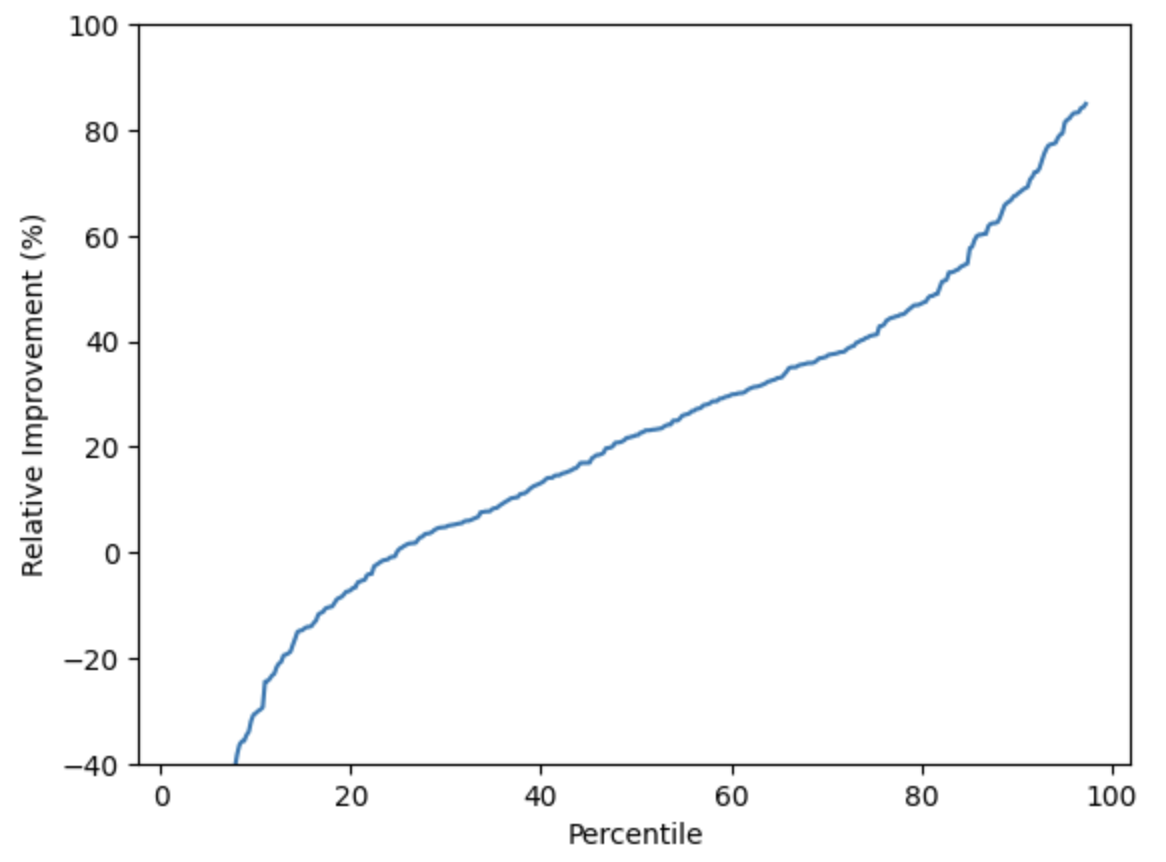}
        \caption{Relative Improvement of Stability Score}
        \label{fig:rel_improv_stability_score}
    \end{subfigure}
    \caption{Comparison of metrics at log scale}
    \label{fig:comparison_of_metrics}
\end{figure}

We evaluated the performance of our AC-optimized SARI model against the
traditionally trained model on 414 time series in the M4 dataset. Overall,
results show that our custom optimization yields substantially improved
stability, often with minimal accuracy trade-offs, whether measured by
one-step-ahead error (using MSE) or multi-step-ahead error and stability (using
forecast AC Score). Figure \ref{fig:comparison_of_metrics} provides a
comprehensive comparison of the MLE-fitted and AC-optimized models in terms of
their forecast AC score \eqref{eq:total_score}, accuracy score
\eqref{eq:energy_score_time_average} and stability score
\eqref{eq:stability_score_time_average}. Specifically, figures
\ref{fig:log_AC_score}, \ref{fig:log_accuracy_score}, and
\ref{fig:log_stability_score} display the log transformed versions of these
scores at each percentile (log transformation is applied due to the large
variation in original scales across the time series). Figures
\ref{fig:rel_improv_AC_score}, \ref{fig:rel_improv_accuracy_score}, and
\ref{fig:rel_improv_stability_score} show the percentile distribution of
relative improvements achieved by the AC-optimized model over the MLE-fitted
model for each of the three metrics. We quote some statistics of the relative
improvement in Table \ref{tab:model_comparison_score}.

\begin{table}[h]
\centering
\caption{Relative improvement of AC-optimized model over MLE-fitted model}
\label{tab:model_comparison_score}
\begin{tabular}{lcccc}
\toprule
\textbf{Statistic} & \textbf{\shortstack{AC Score \\ Improvement}} & \textbf{\shortstack{Accuracy Score \\ Improvement}} & \textbf{\shortstack{Stability Score \\ Improvement}} \\
\midrule
25\% Percentile & 0.80\% & 0.44\% & 0.26\% \\
50\% Percentile (Median) & 2.54\% & 2.04\% & 22.06\% \\
75\% Percentile & 5.09\% & 4.28\% & 41.06\% \\
\bottomrule
\end{tabular}
\end{table}

\textbf{Enhanced Multi Horizon Accuracy and Stability}. 
The ultimate purpose of our metric is to guide or choose the model that
generates forecasts that
\begin{enumerate}
    \item Gain considerable accuracy for multi-horizon prediction without losing
    too much on one step ahead prediction;
    \item Improve consistency for forecasts targeting the same timestamp.
\end{enumerate}
This has been achieved in the out-of-sample tests. 

To quantify the improvement in forecast stability, for each time series we
compute the variance of the mean forecast trajectories targeting the same future
timestamp across different forecast origins. Concretely, for each forecast
origin we take the ensemble mean across the $k = 30$ sample paths as the point
representative of the predictive distribution. Specifically, for each target
timestamp $t$ in the test set, we collect all ensemble mean forecasts
$\hat{y}_{t|s}$, $t-h\leq s \leq t-1$ and calculate their variance
\begin{equation}\label{eq:vertical_variance}
    s^2_{\text{vertical}}(t) = \frac{1}{h-1}\sum_{s=t-h}^{t-1} (\hat{y}_{t|s} - \bar{\hat{y}})^2,
\end{equation}
where $\bar{\hat{y}}$ is the average of forecasts $\hat{y}_{t|s}$ over $t-h \leq
s \leq t-1$. We then average over all target timestamps $t$. This mean value
measures the variance of vertical forecasts. Lower values indicate that
predictions for the same future point are stable throughout forecast origins. 

Finally we compute the ratio 
\begin{equation}
    \frac{s_{\text{vertical}, AC}^2(t)}{s_{\text{vertical}, MLE}^2}.
\end{equation}
A ratio less than one indicates that the AC-optimized model produces more stable
forecasts for that series. Table~\ref{tab:vertical_variance_ratio} reports the
distribution of this per-series ratio across all test series.

\begin{table}[h]
\centering
\caption{Distribution of per-series ratio of mean vertical variance: AC-optimized vs MLE-fitted model.}
\label{tab:vertical_variance_ratio}
\begin{tabular}{lc}
\toprule
\textbf{Statistic} & \textbf{Variance Ratio (AC / MLE)} \\
\midrule
25th Percentile & 73.0\% \\
Median          & 84.2\% \\
75th Percentile & 95.0\% \\
\bottomrule
\end{tabular}
\end{table}

The median ratio of 84.2\% indicates that, for a typical series, the
AC-optimized model reduces the vertical variance of forecast means to roughly
five-sixths of that of the MLE-fitted model. The 25th percentile ratio of 73.0\%
shows that for a quarter of the series the improvement is more pronounced, with
the AC-optimized model reducing vertical variance to below three-quarters of the
MLE baseline. The 75th percentile of 95.0\% indicates that even in less
favorable cases the AC-optimized model still achieves a modest stability gain.
Taken together, stability improvements are consistent across the dataset (though
the magnitude is moderate), reflecting the homoscedastic nature of the SARI
model in which predictive spread with a fixed look-ahead does not vary across
origins.
\begin{remark}
    The modest magnitude of the stability improvement is partly attributable to
    Monte Carlo noise in the ensemble mean. Each sample path carries an
    innovation noise of scale $\sigma$, and so the ensemble mean has residual
    noise of scale $\sigma / \sqrt{k}$. With $k = 30$ sample paths, this
    residual noise is non-negligible and can obscure the smoothing effect that
    AC score optimization imparts on the underlying forecast trajectory. In the
    limit as $k \to \infty$, the ensemble mean converges to the conditional
    expectation and the full stabilizing effect of the AC objective would be
    revealed. The use of a small ensemble is therefore a conservative setting
    for measuring stability gains, and the improvements reported here should be
    interpreted as a lower bound on what the metric can achieve with larger
    sample sizes.
\end{remark}

\textbf{Accuracy Trade-off at Different Horizons}.
One naturally expects a trade-off between the accuracy and the stability of the
forecasts. We measure the trade-off from the following angles:
\begin{enumerate}
    \item Whether improving stability comes at the cost of forecast accuracy. 
    \item Whether aiming at multi-step-ahead comes at the cost of accuracy at
    one-step-ahead.
\end{enumerate}
To evaluate these trade-offs comprehensively, we employ two complementary
evaluation strategies. We begin by assessing performance using our forecast AC
score, as it directly reflects the optimization objective and both the accuracy
and stability of multi-step ahead. We then supplement this with horizon-specific
MSE analysis for two reasons: (1) MSE provides an intuitive, scale-independent
accuracy measure familiar to most practitioners and allows for a more
straightforward interpretation; (2) examining individual horizons allows us to
demonstrate that our multi-horizon optimization does not drastically sacrifice
one-step-ahead forecast quality, which is still an important objective for most
forecasting work. 

The AC-optimized model achieved an improved forecast AC score for 84\% of the
tested series, with significantly improved stability as shown in Figure
\ref{fig:rel_improv_stability_score}. In exchange for the stability improvement,
the AC-optimized model generated slightly less accurate forecasts for
approximately 20\% of the time series (the first quantile in Figure
\ref{fig:rel_improv_accuracy_score}). However, the combined AC score still
improved for the majority of series, demonstrating that stability gains
outweighed the modest accuracy losses for these cases.

\begin{figure}
    \centering
    \includegraphics[width=0.8\linewidth]{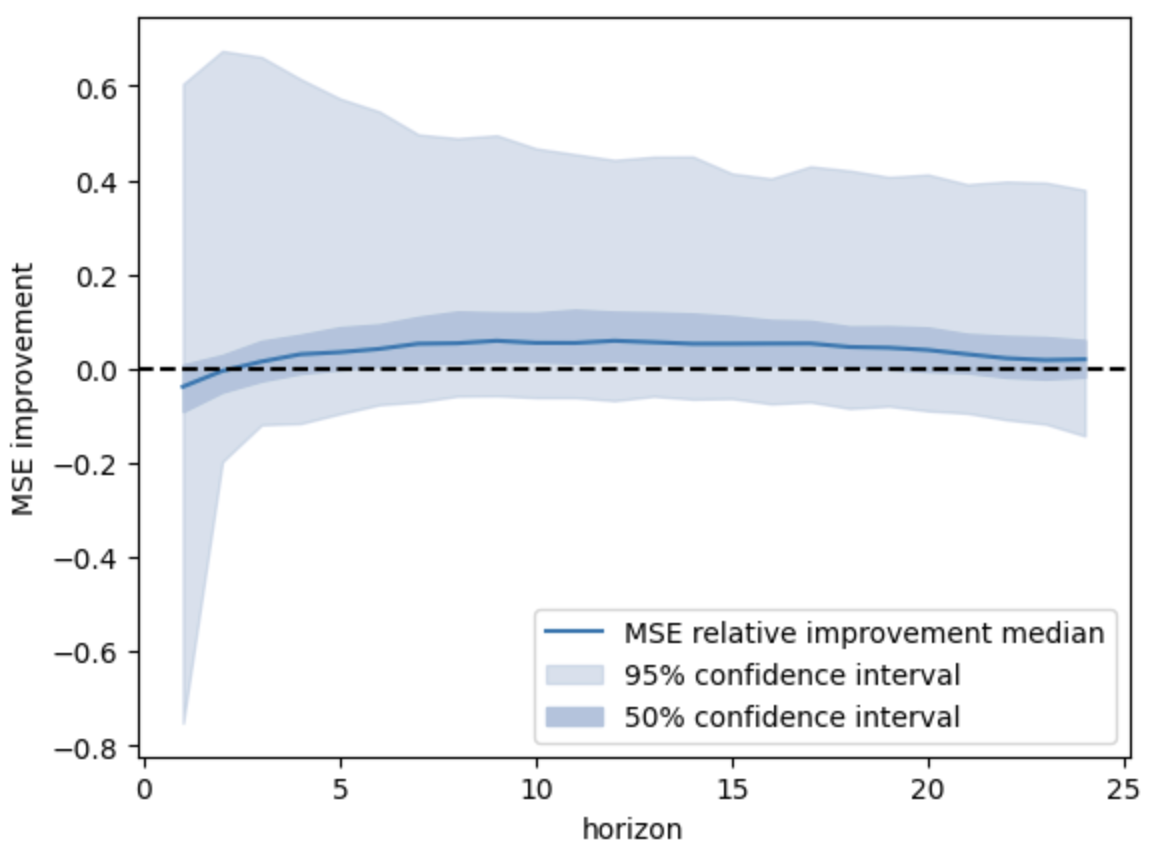}
    \caption{Percentage MSE improvement of the AC-optimized model over the traditional MLE-fitted model.}
    \label{fig:horizon_vs_mse_improvement}
\end{figure}

Next we measure the trade-off between the one-step-ahead accuracy and
multi-step-ahead accuracy plus stability by evaluating the MSE of the
one-step-ahead forecasts. Figure \ref{fig:horizon_vs_mse_improvement} presents
the percentage MSE improvement of the AC-optimized model relative to the
traditional MLE-fitted model across forecast horizons, along with 50\%
confidence intervals. The comparison is evaluated with the percentage
improvement in terms of MSE, as in equation \eqref{eq:relative_MSE_improvement}
\begin{equation}\label{eq:relative_MSE_improvement}
    \text{\% Improvement} = \frac{\text{MSE}_{MLE} - \text{MSE}_{AC}}{\text{MSE}_{MLE}}.
\end{equation}
At the one-step-ahead horizon, our AC-optimized model produces moderately less
accurate forecasts, with its MSE 3.9\% worse compared to that of the traditional
model at the 50th percentile. However, this short-term accuracy cost is quickly
recovered: starting at a horizon 3, our model consistently outperforms the
traditional approach in both accuracy and vertical stability. The improvement
peaks at approximately 6.0\% better MSE around horizons 9--12, then gradually
tapers while remaining positive through horizon 24. This pattern demonstrates
that, while traditional MLE training excels at optimizing one-step-ahead
predictions, it fails to generalize as effectively to longer horizons. In
contrast, our multi-horizon approach generates more accurate medium-to-long
horizon forecasts with only a modest loss of one-step-ahead accuracy. This
trade-off can prove useful for practical decision-making where medium and long
term planning decisions are typically important.

\begin{remark}
    One may notice that \eqref{eq:relative_MSE_improvement} peaks around
    horizons 9--12 and then gradually decreases. This is due to our choice of
    horizon discounting weights, which are linear in the horizon and assign more
    importance to shorter horizons and decay toward zero for longer horizons.
    The improvement remains positive, indicating that the AC-optimized model
    still outperforms the MLE-fitted model to a lesser extent at these horizons.
\end{remark}

\begin{figure}
    \centering
    \includegraphics[width=0.8\linewidth]{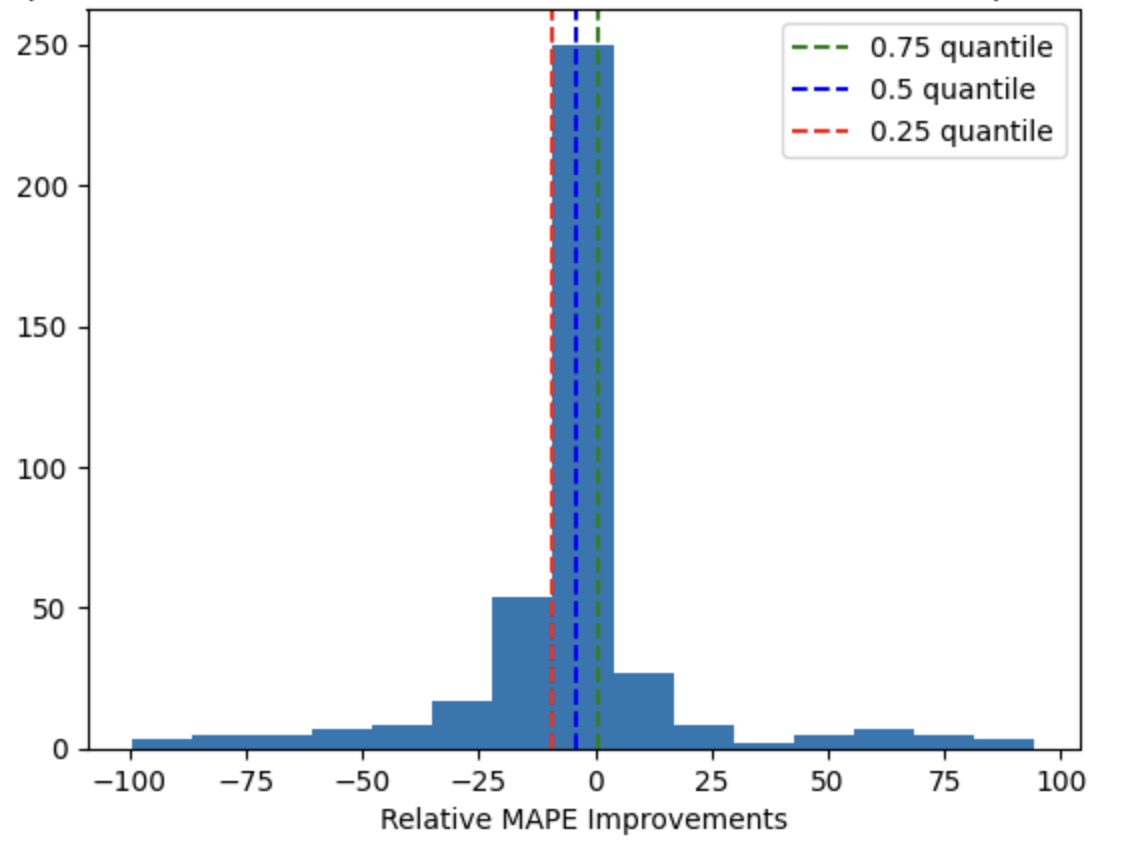}
    \caption{Percentage forecast MSE improvement (\%) of MLE-fitted model relative to AC-optimized model at one step ahead.}
    \label{fig:one_step_ahead_mse_improvement_distribution}
\end{figure}

\textbf{One-Step-Ahead Accuracy Distribution.}
At one step ahead, the AC-optimized model trades off a modest amount of accuracy
in exchange for longer horizon accuracy and stability. To provide a more
detailed picture of this trade-off, we analyze the distribution of performance
differences between the two approaches. Figure
\ref{fig:one_step_ahead_mse_improvement_distribution} is a histogram of the
relative MSE improvement (evaluated as in \eqref{eq:relative_MSE_improvement})
across all test series for one-step-ahead forecasts. The distribution reveals
that, while the traditional model generally performs better at this horizon, the
magnitude of improvement varies considerably across series. The median
improvement (50th percentile) is approximately -3.9\%, indicated by the middle
vertical line, suggesting that for a typical series, the traditional model
achieves modestly better one-step-ahead accuracy. The 25th percentile is
approximately $-9.0\%$, meaning a quarter of series experience up to a 9\% MSE
degradation. The 75th percentile is approximately $+0.8\%$, indicating that for
25\% of series the AC-optimized model achieves comparable or better
one-step-ahead accuracy.

\subsection{Sensitivity to Time Discounting Weights}
We complement our discussion of weight functions in Section
\ref{sec:time_discounting} and assess their influence on optimizing
multi-horizon accuracy. We repeat the experiments using three additional weight
types: uniform ($w(h) = 1$), exponential ($w(h) = e^{-\frac{5}{24}h}$), and
hyperbolic ($w(h) = \frac{1}{1+h}$), each normalized to sum up to 1. 

\begin{figure}
    \centering
    \includegraphics[width=0.8\linewidth]{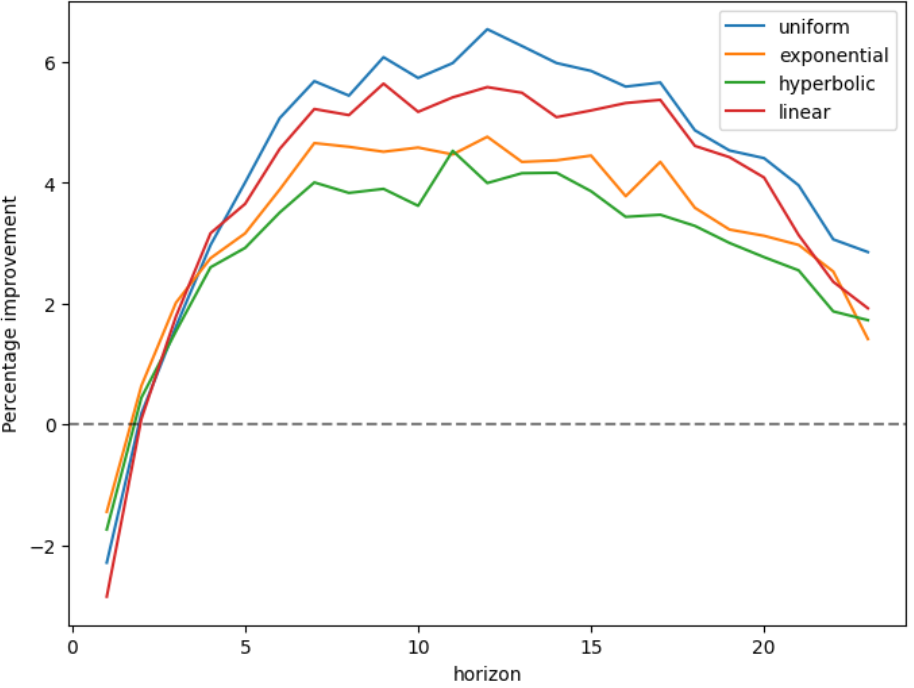}
    \caption{Effect of weight on optimizing multi-horizon accuracy}
    \label{fig:weight_comparison_vertical_stability}
\end{figure}

Figure \ref{fig:weight_comparison_vertical_stability} shows the accuracy
improvement of AC-optimized forecasts over MLE-fitted forecasts, represented by
MSE and separated by weight types. While all schemes trade accuracy at short
horizons for that at long horizons, uniform weights produced the most accurate
forecasts at almost every percentile, followed by linear weights, then
exponential and hyperbolic weights. This result is consistent with the heuristic
that uniform weight assigns equal weights to short and long horizons, coercing
the model to pay attention to the accuracy at these horizons. The other weights
decay for longer horizons, with the rate of decay from slowest to fastest:
linear, hyperbolic, exponential. Setting the rate of decay to approach infinity
will result in the weight converging to $w(h) = \delta(h=1)$ after
normalization.  This limit case is exactly equivalent to training the model
based on one-step-ahead performance. Hence as the weight decays faster from
uniform to exponential, the MSE improvement curves gradually flatten and will
approach the horizontal line $y=0$ as the rate of decay approaches infinity.

\section{Conclusion}

In this paper, we introduced a novel evaluation score, the forecast accuracy and
coherence (AC) score, for multi-horizon time series forecasting that accounts
for both forecast accuracy and stability across prediction horizons. Unlike
traditional metrics that focus solely on point forecast accuracy at individual
horizons, our metric provides an assessment of forecast quality by
simultaneously measuring how well predictions match observed values and how
consistently models predict the same future events from different forecast
origins.

Our metric is built upon the energy score and energy distance. It is highly
customizable, as one can specify relative weights to discount forecasts at
longer horizons based on problem-specific needs, and can adjust the relative
importance of stability with respect to accuracy through the stability
multiplier parameter. This makes the metric applicable across diverse
forecasting contexts, from applications where stability is paramount (such as
capacity planning and resource allocation) to those where accuracy at specific
horizons takes precedence.

To validate the utility of the forecast AC score, we implemented it as a
differentiable objective function for training SARI models. We performed
experiments on the M4 Hourly benchmark dataset. As a result, models trained
using our metric yield consistent improvements over traditionally trained models
in terms of multi-horizon accuracy and stability. In terms of stability, the
AC-optimized model generated out-of-sample forecasts with a median vertical
variance ratio of 84.2\% relative to that of MLE-fitted model, indicating
enhanced forecast consistency. On the accuracy side, the accuracy of forecasts
targeting middle to long horizons are improved for the AC-optimized model. While
the one-step-ahead forecasts suffered a modest 3.9\% degradation in MSE,
forecasts from horizon 3 onward experienced improved accuracy, peaking at
approximately 6\% better MSE around horizons 9--12. This indicates that our
metric is capable of training models that produce more stable and accurate
multi-step forecasts at relatively small cost of the one-step-ahead forecasts.

\section{Future Directions}

In the final section of the paper, we discuss some potential directions for
extending our current work.

\subsection{Application to more Advanced Models}
Our current implementation excludes moving average (MA) terms from the SARIMA
specification. This restriction addresses a computational challenge: in a
differentiable implementation, MA components require forecast error history,
which depends on forecasts from previous training epochs. This creates deep
computational graphs that significantly increase memory requirements and
training time. Future work will incorporate MA components through techniques
such as error approximations to reduce computational cost.

We additionally plan to evaluate the AC score on modern forecasting models
including DeepAR, Temporal Fusion Transformers, and N-BEATS. These experiments
will provide a full picture of the metric's applicability beyond classical
statistical models.

\subsection{Adjusting for Natural Variance Shrinkage}
It is well known that forecast variance typically shrinks naturally as the
horizon decreases and the target time approaches. To isolate true instability
from this natural variance reduction, one can account for the expected evolution
of uncertainty along the path and remove its contribution when computing the
stability term. This ensures that the stability metric captures unexpected
fluctuations rather than predictable variance shrinkage.

\subsection{More Complex Forecasting Architectures} While we have demonstrated
the utility of our metric with the SARI model, the metric can also be utilized
as an objective function in training more complex models.

\subsection{Adjusting Penalties for ``Justified" Revisions }
While forecast stability is desirable in many applications, accuracy typically
takes precedence when these objectives conflict. Stability is primarily valued
as a tie-breaker when comparing forecasts of similar accuracy levels. For
example, consider forecasts for a temperature-sensitive product such as winter
apparel. A sudden temperature drop increases demand, prompting the forecaster to
revise predictions upward. Although this revision is subject to a stability
penalty, it is well-justified because it improves forecast accuracy by
incorporating relevant new information. Our current metric penalizes all
forecast revision equally, regardless of whether the revisions improve accuracy.
Future work could refine the stability penalty to distinguish between
``justified revisions" that genuinely improve forecast accuracy and ``gratuitous
revisions" that do not enhance predictive performance. This extension would
reward forecasters for appropriately updating their predictions in response to
new information while still penalizing erratic behavior that reflects model
instability rather than genuine learning from data. One approach might involve
calculating an ``optimal revision" that achieves the same accuracy as the
current revision, but at minimal instability, and then assigning a penalty based
on the deviation from this optimal revision.

\bibliographystyle{amsplain}
\bibliography{references}

\end{document}